\documentclass[sigconf]{acmart}
\setcopyright{none}
\renewcommand\footnotetextcopyrightpermission[1]{} 
\usepackage{multirow} 
\usepackage{graphicx}
\usepackage{enumitem}
\usepackage{color,soul}
\AtBeginDocument{%
  \providecommand\BibTeX{{%
    \normalfont B\kern-0.5em{\scshape i\kern-0.25em b}\kern-0.8em\TeX}}}
\begin{document}

\title{Spatial-Temporal Synchronous Graph Transformer network (STSGT) for COVID-19 forecasting}

\author{Soumyanil Banerjee}
\affiliation{%
  \department{Department of Computer Science}
  \institution{Wayne State University}
  \city{Detroit, MI}
  \country{USA}}
\email{s.banerjee@wayne.edu}

\author{Ming Dong}
\authornote{Corresponding author}
\affiliation{%
  \department{Department of Computer Science}
  \institution{Wayne State University}
  \city{Detroit, MI}
  \country{USA}}
\email{mdong@wayne.edu}

\author{Weisong Shi}
\affiliation{%
  \department{Department of Computer Science}
  \institution{Wayne State University}
  \city{Detroit, MI}
  \country{USA}}
\email{weisong@wayne.edu}

\renewcommand{\shortauthors}{Banerjee, et al.}

\begin{abstract}
COVID-19 has become a matter of serious concern over the last few years. It has adversely affected numerous people around the globe and has led to the loss of billions of dollars of business capital. In this paper, we propose a novel Spatial-Temporal Synchronous Graph Transformer network (STSGT) to capture the complex spatial and temporal dependency of the COVID-19 time series data and forecast the future status of an evolving pandemic. The layers of STSGT combine the graph convolution network (GCN) with the self-attention mechanism of transformers on a synchronous spatial-temporal graph to capture the dynamically changing pattern of the COVID time series. The spatial-temporal synchronous graph simultaneously captures the spatial and temporal dependencies between the vertices of the graph at a given and subsequent time-steps, which helps capture the heterogeneity in the time series and improve the forecasting accuracy. Our extensive experiments on two publicly available real-world COVID-19 time series datasets demonstrate that STSGT significantly outperforms state-of-the-art algorithms that were designed for spatial-temporal forecasting tasks. Specifically, on average over a 12-day horizon, we observe a potential improvement of 12.19\% and 3.42\% in Mean Absolute Error (MAE) over the next best algorithm while forecasting the daily infected and death cases respectively for the 50 states of US and Washington, D.C. Additionally, STSGT also outperformed others when forecasting the daily infected cases at the state level, e.g., for all the counties in the State of Michigan. The code and models are publicly available at \url{https://github.com/soumbane/STSGT}.
\end{abstract}


\maketitle

\section{Introduction}
The Coronavirus disease 2019 (COVID-19) is caused by severe acute respiratory syndrome coronavirus 2 (SARS-CoV-2). The first known case was identified in Wuhan, China in December 2019, and it has since rapidly spread around the globe. The coronavirus causes infection to the upper respiratory tract which could prove fatal if the immune system does not respond \cite{HARAPAN2020667}. The World Health Organization (WHO) declared COVID-19 as a global pandemic in March, 2020 \cite{Cucinotta2020-sd}. Since then, COVID has disrupted many businesses and households worldwide. Millions of people around the globe got infected with the virus and numerous families lost their loved ones \cite{COVID_Deaths} due to complications that aroused from the virus. Small businesses have been the ones that were the most adversely affected due to COVID \cite{Bartik-COVID-business}. The global economy and financial markets were also severely affected due to COVID \cite{COVID_Economy}. This in turn affected a lot of people worldwide who lost their jobs due to the economic downturn.

Deep learning has revolutionized the areas of computer vision, natural language processing, robotics, medical image processing and many others. Deep learning removes the feature engineering process of traditional machine learning and learns in an end-to-end manner by automatically extracting meaningful and important patterns from the input data. For example, many researchers have used deep learning models for detecting COVID-19 from medical images such as X-rays \cite{MINAEE2020101794, Awan2021_qz}. However, such methods could only be applied for persons already affected by COVID and these are not ideal for prevention of COVID-19. 

In such situations, the prevention of COVID-19 becomes a matter of serious concern as a lot of lives could be saved if we could prevent the disease. The most important question that arises regarding the prevention of COVID is whether we can forecast the future status of the disease. In other words, if we could forecast the total number of infected and death cases in advance, then a lot of precautions could be taken to stop the disease from spreading. This leads us to the problem of time series analysis of COVID cases, where we make use of the past historical COVID data to forecast the future COVID cases. It is also equally important to forecast the COVID cases over the long-term since this would provide some time for the health care workers to prepare for difficult situations and help them serve the infected patients in a more efficient way. Additionally, a long-term forecasting would enable the common people to take safety precautions and prevent the disease. 

Graph Neural Networks (GNNs) are a special type of machine learning algorithms that considers a graph as input and learns to predict the attributes of the vertices and edges of the graph. Recently, GNNs have been used for numerous applications where the data could be modeled as a graph. GNNs could also be combined with time series analysis algorithms to construct the spatial-temporal graph neural network. One of the classic cases where a spatial-temporal graph neural network has been used is traffic forecasting. In traffic forecasting, many researchers have proposed complex spatial-temporal graph neural networks to accurately capture the complex pattern in the spatial-temporal time series data. However, there are very few research works that deal with time series forecasting with COVID data. Some researchers have used spatial-temporal graph neural networks to forecast the COVID cases \cite{kapoor2020examining} but such methods were applied during the start of the pandemic (summer 2020) and as such the size of the dataset was too small to draw an inference. Hence, it becomes critically important that we develop a deep learning model which uses the COVID data from the start of the pandemic to recent times, so that this model could be used for prevention of the disease more effectively.

In this paper, we develop a novel Spatial-Temporal Synchronous Graph Transformer network (STSGT) which uses the historical COVID data to forecast the future infected and death cases for all the US states. In STSGT, first we denote the US states as vertices of a graph and the daily infected or death cases as the vertex features for a given time-step. The physical distances between the states are used as an adjacency matrix. Then, a large spatial-temporal synchronous graph is generated from the historical time-steps, where each day denote a time-step. This novel process ensures that we can capture the complex non-linear spatial-temporal dependencies between adjacent time-steps in a synchronous fashion. We apply the multi-head self-attention mechanism of transformers \cite{vaswani2017attention} on the spatial-temporal synchronous graph to infer the dependency of a vertex of the graph with another vertex. After each vertex finds its importance to all other vertices based on attention scores, we use the graph convolution network (GCN) \cite{kipf2017semi} to aggregate the features (infected or death cases) from neighboring vertices and update the current vertex of the graph. Finally, we forecast the future COVID cases for both short-term, i.e., next day forecasting, and long-term, i.e., forecasting over the next 12 days.

The main contributions of our work are summarized as follows:
\begin{itemize}
  \item We propose a novel Spatial-Temporal Synchronous Graph Transformer network (STSGT) which can simultaneously capture the complex spatial and temporal dependencies of the COVID time series data and accurately forecast the future COVID cases by utilizing the multi-head self-attention mechanism of transformers combined with a graph convolution network (GCN).
  \item To the best of our knowledge, this is the first work that performs such a thorough and detailed analysis with real world COVID-19 time series datasets over a long period of time from the start of the pandemic in March 2020 to Nov 2021 (right before the start of the Omicron wave).
  \item Extensive experiments on two real-world public datasets demonstrate that our model STSGT could accurately forecast the daily infected and death cases at both the national (for all the US states) and state (e.g., for all the counties of the state of Michigan) levels. Our model significantly outperforms other state-of-the-art models designed for spatial-temporal forecasting tasks.
\end{itemize}

The rest of the paper is organized as follows: Section 2 reviews the relevant research works and their applications. Section 3 describes the details of our proposed model STSGT along with its components. Section 4 describes the experimental setup and the results of our experiments. Section 5 presents discussion and future work. Lastly, Section 6 presents our conclusion.

\section{Related Works}
\subsection{Graph Neural Networks}
Graph Neural Networks (GNNs) have been significantly useful when the input data does not have a fixed structure as in an image or text and could be represented as a graph with a set of vertices, edges and an adjacency matrix. GNNs and its variants have been employed for numerous applications such as recommendation systems \cite{wu2020graph}, protein interface prediction \cite{fout2017protein} and semantic segmentation \cite{qi20173d}. For example, a combination of Convolution Neural Networks (CNNs) and graph relation networks have been used for brain connectivity network analysis \cite{banerjee2020deep}.

Graph Convolution Network (GCN) \cite{kipf2017semi} was first developed in a transductive setting to apply the convolutional operation on graphs in the spectral domain. The attention mechanism was introduced in the spatial domain into the feature aggregation process of GCNs by Graph Attention Network (GAT) \cite{velickovic2018graph}, which learns the weights to aggregate features from the neighboring vertices. The first application of graph representation learning to large-scale graphs in an inductive setting was introduced by GraphSAGE \cite{graphsage_2017_hamilton}. Cluster-GCN \cite{cluster_GCN} removed the redundant computation in the neighborhood sampling of GraphSAGE by clustering the graph and then train a GCN. A major limitation of increasing the number of layers of GCN is that it leads to over-smoothing, where vertex representations become indistinguishable as the number of layers of the graph network increases \cite{deep_GCN,li2018deeper,xu2018representation,rong2019dropedge,zhao2019pairnorm}.

\subsection{Spatial-Temporal Forecasting}
Spatial-Temporal forecasting is a fundamentally important problem and has lot of useful applications such as weather forecasting \cite{tekin2021spatio} and action recognition \cite{yan2018spatial}. In smart transportation, traffic forecasting is one of the most challenging problems where the goal is to forecast the future traffic status based on historical observations. Spatial-temporal graph neural networks have been extensively used for traffic forecasting. Spatio-temporal graph convolution networks (STGCN) \cite{yu2018spatio} used a GCN and a gated 1-D CNN to capture spatial and temporal dependencies respectively. Diffusion convolutional recurrent neural networks (DCRNN) \cite{li2018dcrnn_traffic} used the diffusion convolution and the diffusion convolution gated recurrent unit to model the spatial and temporal dependencies respectively. An attention-based approach has been combined with GCN and used with the weekly, daily and hourly traffic data in ASTGCN \cite{guo2019attention}. Graph WaveNet \cite{ijcai_GWNET} uses a gated temporal convolutional network (TCN) \cite{lea2016temporal} with a dilation factor and a GCN for capturing the long-range temporal and spatial dependencies respectively for traffic forecasting. Spatial-temporal synchronous graph convolutional networks (STSGCN) \cite{song2020spatial} applied the GCN on a synchronous spatial-temporal graph to capture the complex spatial-temporal dependencies and heterogeneity in traffic data. Spatial-temporal fusion graph neural networks (STFGNN) \cite{li2021spatial} fused a temporal graph with the spatial graph to forecast the future traffic conditions. Spatial-temporal graph neural networks has also recently been used to forecast the COVID-19 cases \cite{kapoor2020examining}. 

\subsection{Transformers}
One of the most prominent architectures that have revolutionized the Natural Language Processing (NLP) field are transformers \cite{vaswani2017attention}. Transformers introduced the multi-head self-attention mechanism which could capture the relationship between every pair of words in a sentence in a parallel manner, thereby improving over the performance of serial architectures such as long short-term memory (LSTM) \cite{hochreiter1997long} and gated recurrent unit (GRU) \cite{chung2014empirical}, which have the vanishing gradient problem \cite{hochreiter1998vanishing}. Transformers and its recent variants such as BERT \cite{devlin2018bert} and large language models such as GPT-3 \cite{brown2020language} have been used for numerous NLP applications such as machine translation, sentiment classification, speech recognition and many others. 

The self-attention mechanism of transformers have also been used in computer vision by vision transformers \cite{dosovitskiy2020image} for image recognition at scale. Recently, the self-attention mechanism of transformers has been adopted to the spatial domain of GNNs with graphormer \cite{ying2021transformers} and to the spatial-temporal forecasting domain with the spatial-temporal transformer network (STTN) \cite{xu2020spatial}.   

\section{Our Approach}
Our proposed Spatial-Temporal Synchronous Graph Transformer network (STSGT) could be summarized with three main points: 1) Generate the samples from the original time series data with a sliding window approach; 2) For each sample, connect each vertex with itself at the previous and the next time steps to construct a spatial-temporal synchronous graph; 3) Use STSGT layers to capture the complex spatial-temporal correlations and then forecast the future status of the spatial-temporal graph network. 

In STSGT, we generate a large synchronous spatial-temporal graph with several days historical data and then calculate the multi-head self-attention between every pair of vertices of this large graph. This novel process not only captures the relationship between the COVID infected or death cases between neighboring states or counties in a given day but also simultaneously captures the relationship between those states or counties in the previous and following days. Note that STSGT is fundamentally different from existing approaches for spatial-temporal forecasting. First, in most other graph neural network-based spatial-temporal models, a GCN is used to capture the spatial dependency for a given time-step, and a separate 1D convolution unit or attention unit is used to capture the temporal dependency between the time-steps, while STSGT used the multi-head self-attentions to capture both spatial and temporal dependency synchronously. In the literature, the most similar model to our approach is STTN \cite{xu2020spatial}, which has separate spatial and temporal transformer units and hence cannot simultaneously capture the complex spatial-temporal dependencies between adjacent time steps. The synchronous mechanism in our model helps us to achieve better performance in forecasting.

In the following sections, we provide a detailed discussion on every component of STSGT.

\subsection{Generate samples using a Sliding Window}
\begin{figure}[htbp]
\centering
\includegraphics[width=0.99\linewidth]{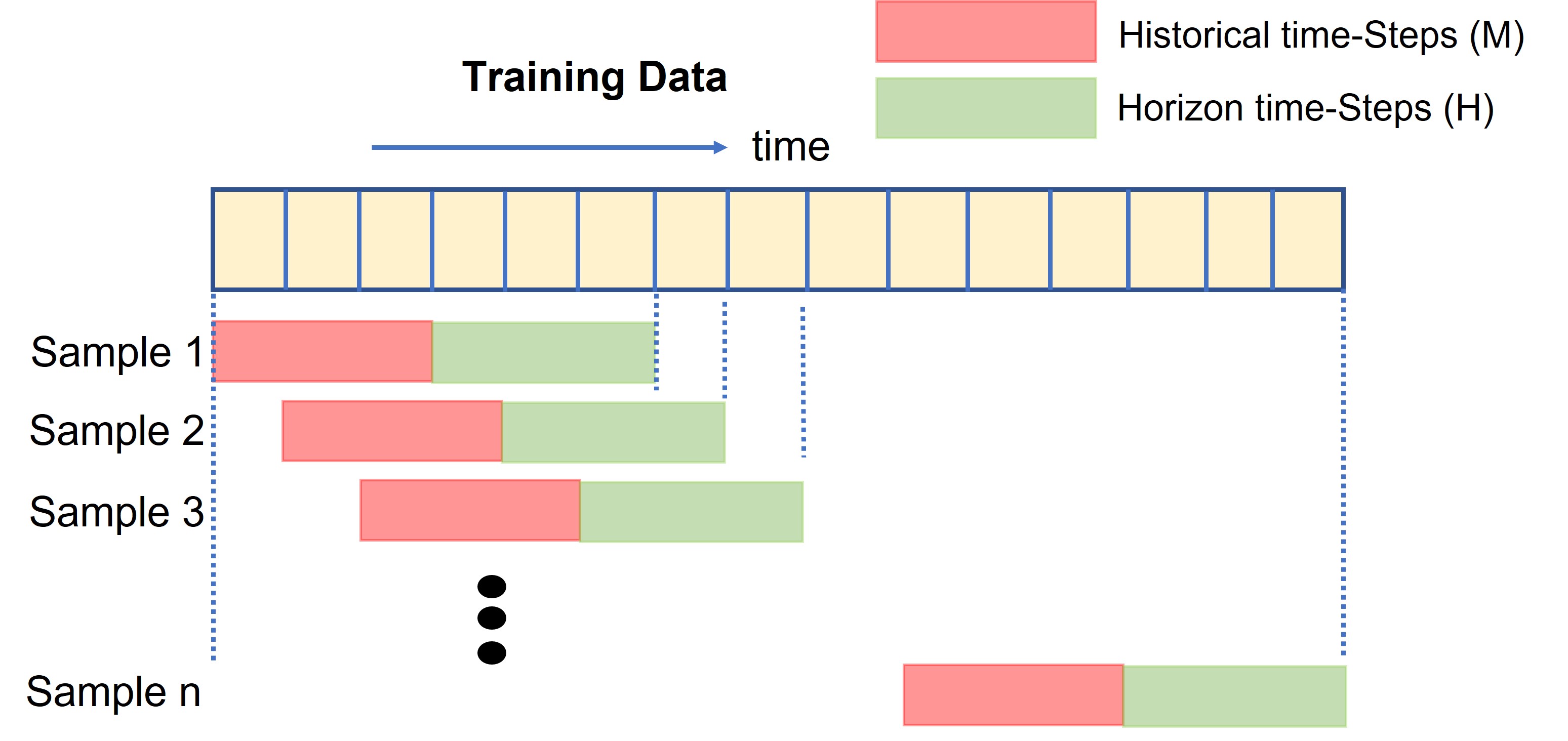}
\caption{Generating the samples from the training data using a sliding window approach.}
\label{fig:figure1}
\end{figure}

We use a sliding window approach to generate the samples from the COVID-19 time series datasets as shown in Figure \ref{fig:figure1}. 

Each of the time step comprises of a spatial graph $G = (V, E, A)$, where $V$ and $E$ denote the set of vertices and edges of $G$ respectively. The number of vertices of the graph is denoted as $N$, where $N = |V|$. ``$A$'' denotes the adjacency matrix of $G$, where $A \in \mathbb{R}^{N \times N}$. For each vertex $N$ of the graph $G$, there are certain features $F$ that are associated with it. Hence, we denote graph signal matrix for each time-step $t$ of the time series as $X_{(t)}$, where $X_{(t)} \in \mathbb{R}^{N \times F}$. 

Then, we can formulate the problem of COVID-19 time series forecasting as learning the function $f$ which uses the historical $M$ observations to forecast the future $H$ observations:
\begin{equation}
    \hat{X}_{(t+1)}, \hat{X}_{(t+2)}, ..., \hat{X}_{(t+H)} = f \left (X_{(t)}, X_{(t-1)}, ..., X_{(t-M+1)} \right)
    \label{equation:formulation}
\end{equation}
where, $\hat{X}$ and $X$ denote the $H$ forecasted and $M$ ground truth observations respectively.

\subsection{Construction of the Spatial-Temporal Synchronous Graph}
\begin{figure}[htbp]
\centering
\includegraphics[width=0.99\linewidth]{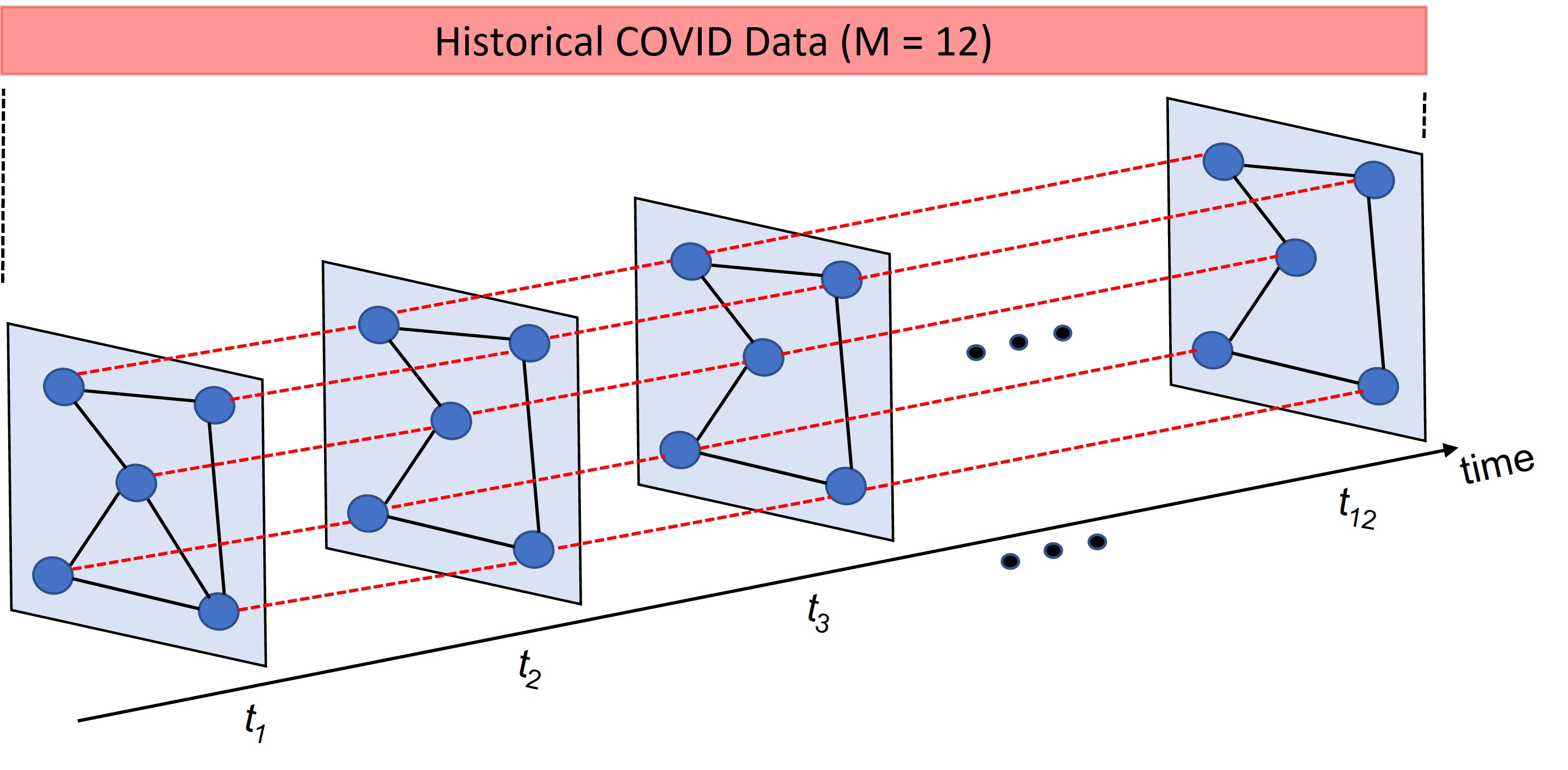}
\caption{Spatial-Temporal Synchronous Graph $\tilde{G}$ generated from the historical $M$ time steps. Here, we have illustrated the case when $M=12$.}
\label{fig:figure2}
\end{figure}

The goal behind the construction of the spatial-temporal synchronous graph is to capture the complex spatial-temporal dependencies that are present between vertices of previous and subsequent time-steps. To achieve this goal, we connected each vertex $N$ of the graph $G$ to itself in the previous and next time-step for the entire time period $M$ as illustrated in Figure \ref{fig:figure2}. This results in a larger spatial-temporal graph $\tilde{G}$ where a given vertex at a particular time-step can aggregate information from another neighborhood vertex of previous and subsequent time-steps. The larger spatial-temporal graph also helps in capturing the long-range dependencies between the vertices of different time-steps.

The adjacency matrix of the spatial graph $G$ with $N$ vertices at a given time-step $t$ is given as $A \in \mathbb{R}^{N \times N}$. Hence, the spatial-temporal synchronous graph $\tilde{G}$ that we construct will have its adjacency matrix $\tilde{A}$, where $\tilde{A} \in \mathbb{R}^{MN \times MN}$. In this large graph $\tilde{G}$, a given vertex's index $p (1 \leq p \leq N)$ in the spatial graph $G$ will be mapped to a new index in the large spatial-temporal graph $\tilde{G}$, and the new index will be: $(t - 1)N + p$, where $t (1 \leq t \leq M)$ is the time-step number in the larger spatial-temporal synchronous graph $\tilde{G}$. 

In the large spatial-temporal synchronous graph $\tilde{G}$, when a given vertex at time-step $t_i$ is connected to itself at subsequent time-step $t_j$, the corresponding value in the large adjacency matrix $\tilde{A}$ with $MN$ vertices is set to be 1. Hence, the adjacency matrix of $\tilde{A}$ is formulated as:
\begin{equation}
    \tilde{A}_{i,j}= 
    \begin{cases}
    1,& \text{if $v_i$ connects to $v_j$}\\
    0,              & \text{otherwise}
    \end{cases}
    \label{equation:sync_adj}
\end{equation}
where $v_i$ and $v_j$ denotes the same vertex $i$ and $j$ respectively in subsequent time-steps of the spatial-temporal synchronous graph $\tilde{G}$. 

\begin{figure}[htbp]
\centering
\includegraphics[width=0.9\linewidth]{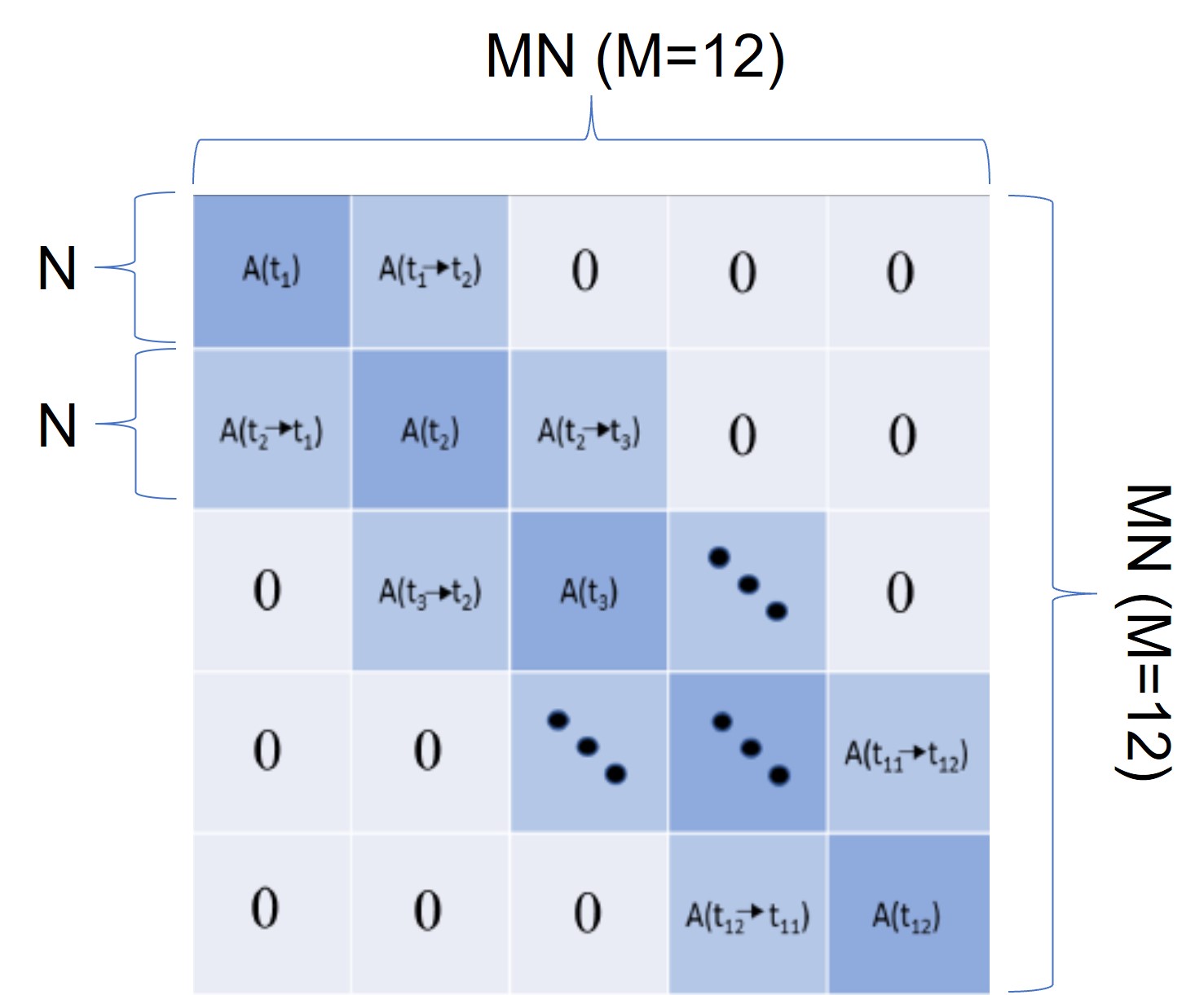}
\caption{Adjacency Matrix $\tilde{A}$ of the Spatial-Temporal Synchronous Graph $\tilde{G}$. Here, we have illustrated the case when $M=12$.}
\label{fig:figure3}
\end{figure}

We have shown the adjacency matrix $\tilde{A}$ of the spatial-temporal synchronous graph $\tilde{G}$ in Figure \ref{fig:figure3}. The diagonal of $\tilde{A}$ are the adjacency matrices $A(t_k)$ of the corresponding spatial graphs $G(t_k)$ for time-steps $t_k$ where $1 \leq k \leq M$. Hence, we can say that $\tilde{A}_{i,i} = A(t_k)$, where $i = (k-1)N$ to $kN$ for $1 \leq k \leq M$.

The off-diagonal matrices of $\tilde{A}$ indicate the connection between each vertex to itself in the previous or subsequent time-steps. Hence, we can say that the upper-triangular side of $\tilde{A}$, i.e. $A(t_k \rightarrow t_{k+1})$ for $1 \leq k \leq (M-1)$ is formed as: $\tilde{A}_{i,j} = 1$, where $i = (k-1)N+p$, $j = kN+p$ and $i \neq j$, for $1 \leq k \leq (M-1)$ and $1 \leq p \leq N$. In a similar manner, the lower triangular side of $\tilde{A}$, i.e. $A(t_{k+1} \rightarrow t_k)$ for $1 \leq k \leq (M-1)$ is symmetric to the upper triangular side $A(t_k \rightarrow t_{k+1})$ for $1 \leq k \leq (M-1)$. All the other elements are zero as indicated in Figure \ref{fig:figure3}. 

In our approach, the spatial adjacency matrix $A$ for graph $G(t_k)$ at time-step $t_k$ is a weighted adjacency matrix formed by the physical distances between the vertices of $G(t_k)$ which we denote as the states of US or the counties in a state. Hence, for this adjacency matrix $A$, the feature aggregation process of GCN in our model's STSGT layer will not be affected as each vertex of $G(t_k)$ has a different influence on its neighboring vertices. But for the adjacency matrix $\tilde{A}$ of the large spatial-temporal synchronous graph $\tilde{G}$, the off-diagonal matrices $A(t_k \rightarrow t_{k+1})$ and $A(t_{k+1} \rightarrow t_k)$ ($1 \leq k \leq (M-1)$) have entries that are $1$. Our goal is to learn the temporal dependency between same vertices in adjacent time-steps ($v_i$ and $v_j$) in an end-to-end fashion. This would make our feature aggregation process more general and flexible since the features between vertices of adjacent time-steps would be aggregated based on the learned temporal dependency. Hence, we define a mask matrix $\tilde{A}_{mask}$ with a similar shape as $\tilde{A}$, i.e. $\tilde{A}_{mask} \in \mathbb{R}^{MN \times MN}$. We calculate the Hadamard-product between the mask matrix and the adjacency matrix $\tilde{A}$ to form an updated matrix as follows:

\begin{equation}
    \tilde{A}_{updated} = \tilde{A}_{mask} \circ \tilde{A}
    \label{equation:mask_mat}
\end{equation}
where the mask matrix $\tilde{A}_{mask}$ contains the learnable weights between the vertices of adjacent time-steps $v_i$ and $v_j$ during the training process, $\circ$ denotes the Hadamard-product and $\tilde{A}_{updated} \in \mathbb{R}^{MN \times MN}$ denotes the updated adjacency matrix. Subsequently, $\tilde{A}_{updated}$ is used for all GCN operations on the spatial-temporal synchronous graph $\tilde{G}$.  

\subsection{Spatial-Temporal Synchronous Graph Transformer Network}
\begin{figure}[htbp]
\centering
\includegraphics[width=0.9\linewidth]{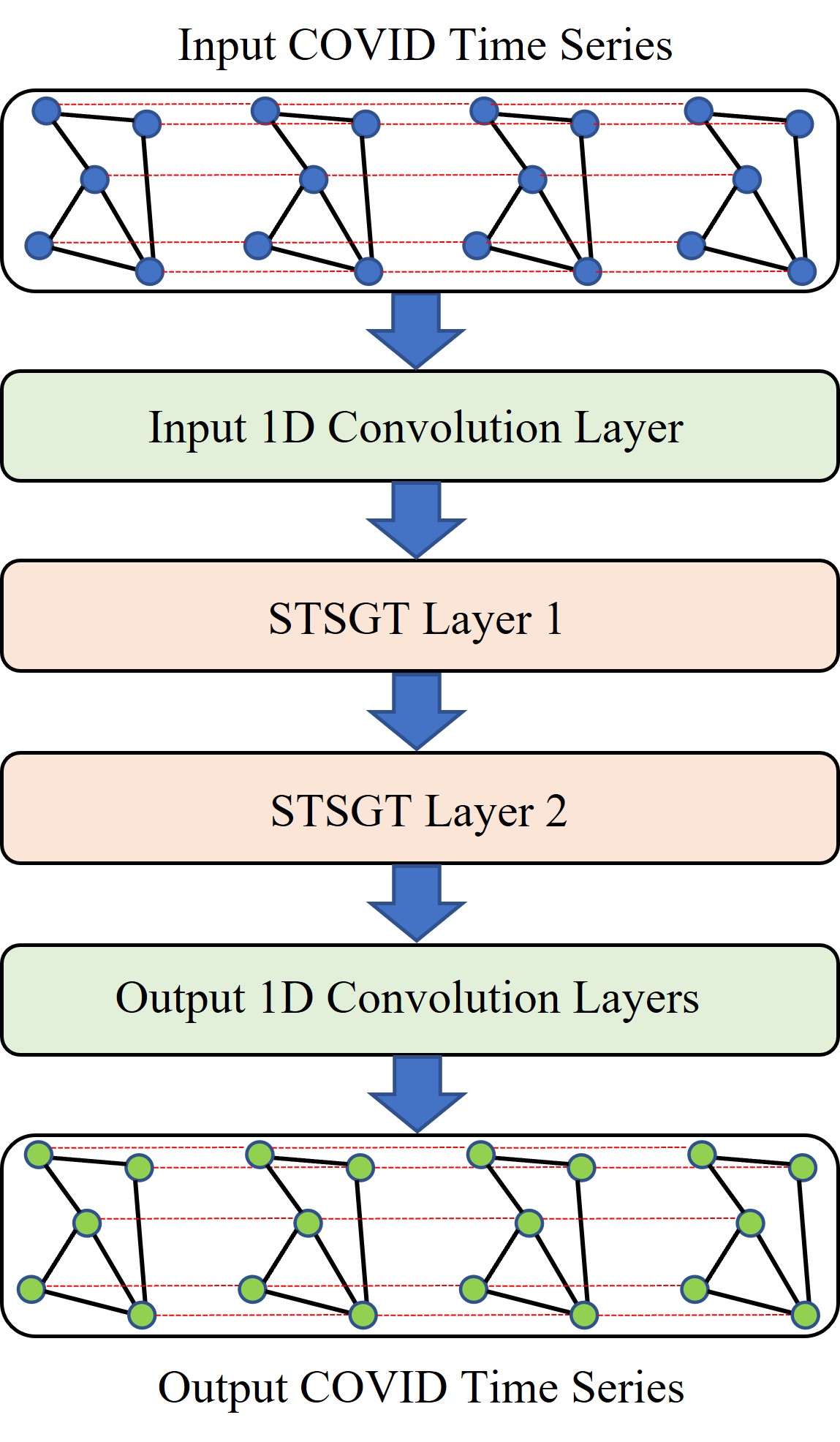}
\caption{The architecture of our proposed Spatial-Temporal Synchronous Graph Transformer network (STSGT).}
\label{fig:figure4}
\end{figure}

The architecture of our proposed Spatial-Temporal Synchronous Transformer network (STSGT) is shown in Figure \ref{fig:figure4}. 

The input of STSGT is generated from the graph signal matrix $X_{(t)} \in \mathbb{R}^{N \times F}$ by considering $M$ continuous time-steps. Hence, we generate the spatial-temporal synchronous graph signal matrix $\tilde{X} \in \mathbb{R}^{M \times N \times F}$ which is basically the graph signal matrix of the large spatial-temporal synchronous graph $\tilde{G}$.

The input graph signal matrix is passed through a single 1-D convolutional layer to map the input to a higher dimensional space. Hence, the output of the 1-D convolutional layer is the spatial-temporal synchronous graph signal matrix $\tilde{X} \in \mathbb{R}^{M \times N \times C_{in}}$ where $C_{in} > F$, and then $\tilde{X}$ is passed on to the STSGT layers.

\subsubsection{Spatial-Temporal Synchronous Graph Transformer layer}
\begin{figure}[htbp]
\centering
\includegraphics[width=\linewidth]{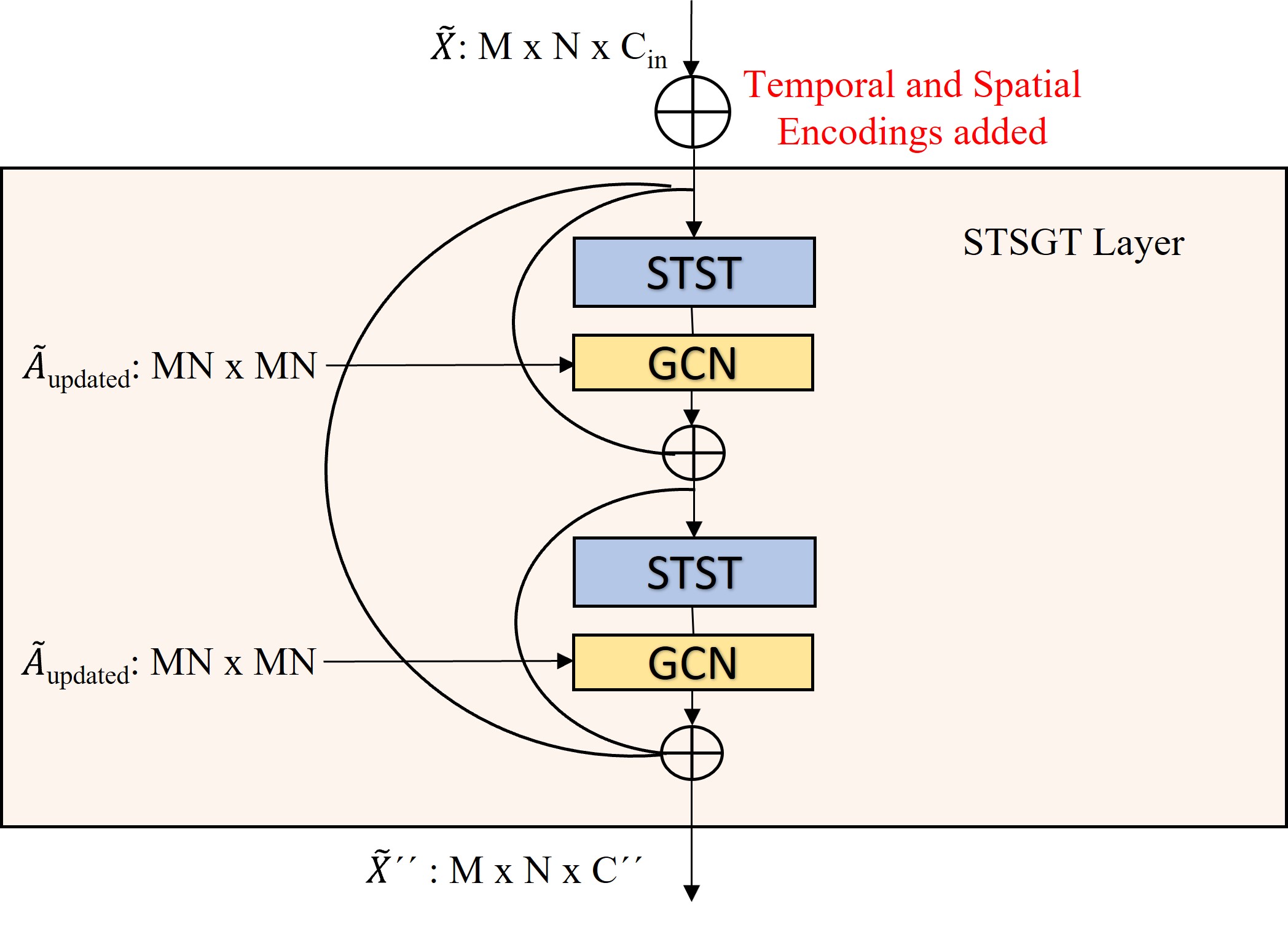}
\caption{The architecture of one Spatial-Temporal Synchronous Graph Transformer layer. The adjacency matrix of the spatial-temporal synchronous graph $\tilde{A}_{updated}$ is used by the GCN.}
\label{fig:figure5}
\end{figure}

\hfill\\
The architecture of the STSGT Layer is illustrated in Figure \ref{fig:figure5}. 

Each layer comprises of the Spatial-Temporal Synchronous Transformer (STST) for calculating the self-attention between the vertices of the spatial-temporal synchronous graph $\tilde{G}$. A conventional GCN follows the STST to aggregate the information obtained through self-attention mechanism, from neighboring vertices of all vertices of $\tilde{X}$. A GCN layer on the output of STST is defined as:
\begin{equation}
    \tilde{X''} = \sigma \left( \tilde{A}_{updated} \tilde{X'} W + b \right)
    \label{equation:gcn_aggr}
\end{equation}
where $\tilde{X'} \in \mathbb{R}^{M \times N \times C'}$ is the output of STST, $\tilde{A}_{updated} \in \mathbb{R}^{MN \times MN}$ is the updated spatial-temporal synchronous adjacency matrix, $W \in \mathbb{R}^{C' \times C''}$ is the learnable weight matrix, $b \in \mathbb{R}^{C''}$ is the learnable bias, $\sigma$ is the non-linearity such as ReLU applied after the graph convolution, and $\tilde{X''} \in \mathbb{R}^{M \times N \times C''}$ is the output of the GCN layer.

Additionally, residual connections are added as shown in Figure \ref{fig:figure5} to allow the gradients to back-propagate to the initial layers of the model during the training process. Finally, the output of a STSGT layer is an updated spatial-temporal graph signal matrix $\tilde{X''} \in \mathbb{R}^{M \times N \times C''}$ which is passed on to the next STSGT layer.

\subsubsection{Temporal and Spatial Encodings}\hfill\\
Temporal and spatial encodings are added to the input spatial-temporal graph signal matrix $\tilde{X} \in \mathbb{R}^{M \times N \times C_{in}}$, before passing it on to the first STSGT layer. This helps the model to distinguish between the different time-steps (temporal encodings) and the different vertices of a given time-step (spatial encodings). These temporal and spatial encodings are added as follows:
\begin{equation}
    \tilde{X} = \tilde{X} + T_{enc} + S_{enc}
    \label{equation:encodings}
\end{equation}
where the graph signal matrix after temporal and spatial encoding is $\tilde{X} \in \mathbb{R}^{M \times N \times C_{in}}$. The temporal encoding added is $T_{enc} \in \mathbb{R}^{M \times 1 \times C_{in}}$ which basically assigns same values for all vertices $N$ of a given time-step $t$ and different values for vertices in different time-steps. The spatial encoding added is $S_{enc} \in \mathbb{R}^{1 \times N \times C_{in}}$ which basically assigns same values for a given vertex $N$ across all time-steps and different values for vertices in the same time-step $t$.


\subsubsection{Spatial-Temporal Synchronous Self-Attention}
\begin{figure}[htbp]
\centering
\includegraphics[width=\linewidth]{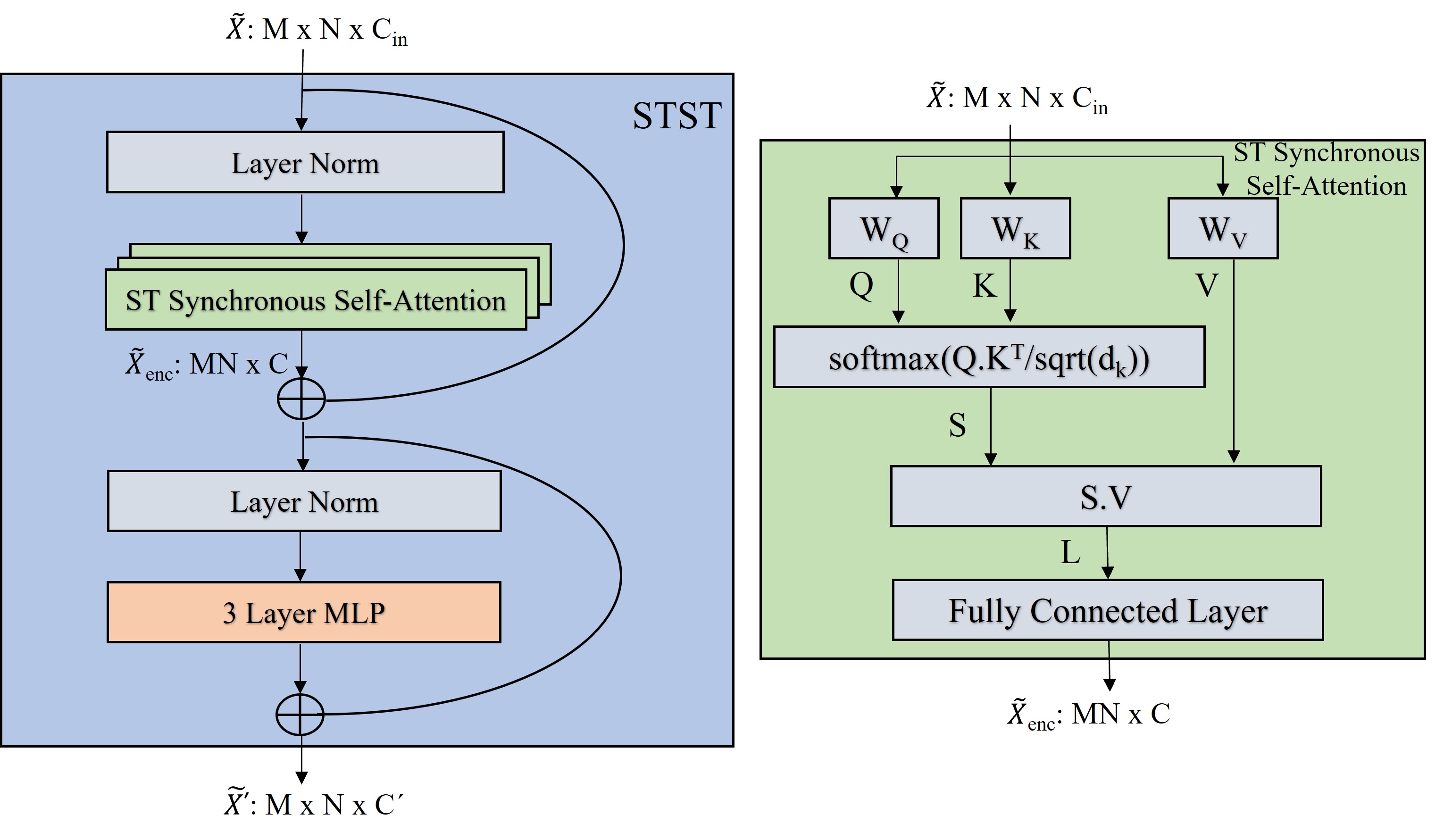}
\caption{(left) The architecture of STST and (right) The architecture of one head of the Spatial-Temporal Synchronous Self-Attention within each STST.}
\label{fig:figure6}
\end{figure}

\hfill\\
The architecture of the Spatial-Temporal Synchronous Transformer (STST) within each STSGT layer is illustrated in Figure \ref{fig:figure6} (left). First, the layer normalization is applied to the input graph signal matrix $\tilde{X} \in \mathbb{R}^{M \times N \times C_{in}}$. Subsequently, the output of the layer norm is passed to a multi-head spatial-temporal synchronous self-attention mechanism as illustrated in Figure \ref{fig:figure6} (right). Layer normalization is applied again to the output of the multi-head self-attention. Finally, a three layer Multi-Layer Perceptron (MLP) is used to output the encoded graph signal matrix $\tilde{X'} \in \mathbb{R}^{M \times N \times C'}$. 

The architecture of each head of the spatial-temporal synchronous self-attention mechanism is illustrated in Figure \ref{fig:figure6} (right). The input to each head is the spatial-temporal graph signal matrix $\tilde{X} \in \mathbb{R}^{M \times N \times C_{in}}$. This matrix is reshaped to $\tilde{X} \in \mathbb{R}^{MN \times C_{in}}$. Then $\tilde{X}$ is projected to three higher dimensional latent subspaces by generating the Query(Q), Key(K) and Value(V) matrices. The Q, K and V matrices are generated as follows:
\begin{equation}
    Q = \tilde{X} W_{Q}
    \qquad
    K = \tilde{X} W_{K}
    \qquad
    V = \tilde{X} W_{V}
    \label{equation:gen_qkv}
\end{equation}
where, $Q \in \mathbb{R}^{MN \times d_{q}}$ and $W_{Q} \in \mathbb{R}^{C_{in} \times d_{q}}$, $K \in \mathbb{R}^{MN \times d_{k}}$ and $W_{K} \in \mathbb{R}^{C_{in} \times d_{k}}$, $V \in \mathbb{R}^{MN \times d_{v}}$ and $W_{V} \in \mathbb{R}^{C_{in} \times d_{v}}$. Here, $d_{q}=d_{k}=d{v}$ and the projection matrices $W_{Q}, W_{K}, W_{V}$ are feed forward neural networks.

In a similar manner as the encoder of the transformer \cite{vaswani2017attention}, we calculate the similarity or attention score $S$ between the query and key matrices in the higher dimensional latent subspace. The attention score $S$ is calculated as:
\begin{equation}
    S = softmax \left( \frac{Q K^{T}}{\sqrt{d_{k}}} \right)
    \label{equation:attn_score}
\end{equation}
where, $S \in \mathbb{R}^{MN \times MN}$ is the attention score obtained from self-attention between all the vertices of the spatial-temporal synchronous graph $\tilde{G}$ with graph signal matrix $\tilde{X}$. $Q$ and $K$ are the query and key matrices respectively, and $d_{k}$ is the dimension of the key matrix $K$. The softmax function normalizes the scores and the scaling factor $\frac{1}{\sqrt{d_{k}}}$ prevents the saturation due to the softmax function. 

Finally, the self-attention scores are multiplied by the value matrix $V$ to update the features of each vertex of $\tilde{X}$. The update is done as follows:
\begin{equation}
    L = S V
    \label{equation:cal_val}
\end{equation}
where $S$ and $V$ are the attention score and value matrix respectively. The matrix $L \in \mathbb{R}^{MN \times d_{v}}$ obtained with equation \ref{equation:cal_val} is passed through a fully connected layer to update the output spatial-temporal synchronous graph signal matrix $\tilde{X}_{enc} \in \mathbb{R}^{MN \times C}$. 

\subsubsection{Forecasting the future COVID status}\hfill\\
We forecast the future values of the spatial-temporal COVID time series with an output layer (see Figure \ref{fig:figure4}) which consists of two 1D convolution layers to map the output of the last STSGT layer to the original dimension of the time series data. Let us denote the output of the last STSGT layer in Figure \ref{fig:figure4} as $\tilde{X} \in \mathbb{R}^{M \times N \times C'}$, which is then transposed and reshaped to $\tilde{X}^{T} \in \mathbb{R}^{N \times M C'}$. We then use two 1D convolution layers on $\tilde{X}^{T}$ as follows:
\begin{equation}
    \hat{X} = \sigma \left( \tilde{X}^{T} W_1 + b_1 \right).W_2 + b_2
    \label{equation:out_conv}
\end{equation}
where, $\hat{X} \in \mathbb{R}^{N \times H}$ is output of our STSGT model, $\sigma$ is the ReLU non-linearity, $W_1 \in \mathbb{R}^{M C'\times C''}$ and $b_1 \in \mathbb{R}^{C''}$ are the learnable weights and bias of the first 1D convolution layer and $W_2 \in \mathbb{R}^{C'' \times H}$ and $b_2 \in \mathbb{R}^{H}$ are the learnable weights and bias of the second 1D convolution layer respectively. Here, $C''$ is the hidden layer dimension of the output layers and $H$ is the length of the forecasting horizon. Finally, the output $\hat{X}$ is transposed to $\hat{X}^{T} \in \mathbb{R}^{H \times N}$. 

It should be noted that we forecast all the $H$ horizon time-steps $\hat{X}_{(t+1)}, \hat{X}_{(t+2)}, ..., \hat{X}_{(t+H)}$ at the same time instead of recursively forecasting one of the horizon time-step at a given time as done in previous works \cite{song2020spatial, li2018dcrnn_traffic, yu2018spatio}.

\section{Experiments}
We evaluated our STSGT framework with three main tasks: 1) forecasting the COVID-19 daily infected cases for the 50 states of US and Washington, D.C.; 2) forecasting the COVID-19 daily infected cases at the state level with 83 counties for the state of Michigan; and 3) forecasting the COVID-19 daily death cases for the 50 states of US and Washington, D.C. We compared the performance of our proposed approach with baselines and state-of-the-art spatial-temporal forecasting algorithms. All our experiments were performed using the PyTorch 1.8.0 framework \cite{NEURIPS2019_9015}. 

\subsection{Datasets}
\subsubsection{John Hopkins University (JHU) COVID-19 Dataset}\hfill\\
We used the publicly available COVID-19 Data Repository by the Center for Systems Science and Engineering (CSSE) at Johns Hopkins University (\url{https://github.com/CSSEGISandData/COVID-19}) \cite{JHU_data_1}. Specifically, we used the time series dataset which is updated regularly with the number of infected and death cases across all states of the US. We used the time series data from March 15, 2020 to Nov 30, 2021 spanning 626 days. 

\subsubsection{New York Times (NYT) COVID-19 Dataset}\hfill\\
We used the publicly available data from The New York Times, based on reports from state and local health agencies \cite{NYT_data}. Similar to the JHU dataset, we used the time series data for the daily infected and death cases across all US states. We used the time series data from March 18, 2020 to Nov 30, 2021 spanning 623 days.

\subsection{Comparison Methods}
\subsubsection{Baseline Method}\hfill\\
We used the following baseline method for spatial-temporal forecasting, to compare with our proposed approach.

\begin{itemize}
    \item \textbf{Auto Regressive Integrated Moving Average (ARIMA)} \cite{li2018dcrnn_traffic} - ARIMA consists of an auto-regressive (AR) and a moving average (MA) component and is used to forecast future values of a stationary time series. For our experiments, we used an order of (5, 1, 0) for ARIMA. That is, we used 5 AR and 0 MA components with first order difference (d=1) on the input time series. We used ARIMA as a baseline comparison method since it is one of the most common algorithms for time-series analysis, and it has been extensively used as a comparison method by several state-of-the-art spatial-temporal forecasting algorithms. 
\end{itemize}

\subsubsection{State-of-the-art Spatial-temporal Forecasting Algorithms}\hfill\\
We compared our proposed approach with the following state-of-the-art methods that were designed for spatial-temporal forecasting. For each of the models below, the number of features or channels for each vertex of the graph is 1, which is the daily infected or death cases. The number of vertices for the graph per time-step is 51 (50 US states and Washington, D.C.) and 83 (for Michigan). The history (M) and horizon (H) for the sliding window is selected as 12 for all the spatial-temporal forecasting algorithms below. The default values are used for the hyper-parameters in each model, which we detailed in their descriptions below.

\begin{itemize}
    \item \textbf{Spatial-Temporal Graph Convolution Network (STGCN)} \cite{yu2018spatio} - STGCN uses the spatial-temporal convolution blocks to forecast the horizon. Each of its spatial-temporal blocks consists of spectral graph convolution to capture the spatial dependency between vertices of same time-step and an independent temporal gated convolution block to capture the temporal dependency between the time-steps. We used the STGCN implementation (\url{https://github.com/FelixOpolka/STGCN-PyTorch}) with 16 spatial and output channels for each STGCN block. 
    
    \item \textbf{Attention-based Spatial-Temporal Graph Convolution Network (ASTGCN(r))} \cite{guo2019attention} - ASTGCN concatenates the output of several spatial-temporal blocks used independently on the hourly, daily and weekly spatial-temporal data to forecast the next hour status. Each of the spatial-temporal blocks consists of spectral graph convolution to capture the spatial dependency and a combination of spatial and temporal attention mechanism to capture the dynamic spatial and temporal correlations of the spatial-temporal data. We used only the recent component of ASTGCN (i.e., ASTGCN(r)), equivalent to only using the hourly data, to have a fair comparison with other methods. We used the ASTGCN implementation (\url{https://github.com/guoshnBJTU/ASTGCN-r-pytorch}) with 2 ASTGCN blocks, 16 time filters with stride 1 and 16 Chebyshev filters of order 2.  
    
    \item \textbf{Graph WaveNet} \cite{ijcai_GWNET} - Graph WaveNet comprises of multiple layers with skip connections where each layer consists of two gated temporal convolutional networks (TCN) \cite{lea2016temporal}. A dilation factor is used for each TCN to consider the long-range temporal dependencies of the time series. The gated TCN is followed by an independent graph convolutional layer to capture the spatial dependencies. We used the Graph WaveNet implementation (\url{https://github.com/nnzhan/Graph-WaveNet}) with the ``double transition'' adjacency matrix. We used 2 blocks with each block having 2 layers. We used 16 residual and dilation channels along with 128 skip channels and 256 end channels. A $(1,2)$ kernel size was used for the filter and gate convolutions. 
    
    \item \textbf{Spatial-Temporal Transformer Network (STTN)} \cite{xu2020spatial} - STTN has several spatial-temporal blocks, and each of these blocks consists of an independent spatial and temporal transformer. The spatial transformer combines self-attention with a GCN using a gated mechanism to capture the spatial dependency. The output of the spatial transformer is passed to the temporal transformer which uses the self-attention mechanism to capture the temporal dependencies. We used the STTN implementation (\url{https://github.com/wubin5/STTN}) with 2 layers of spatial and temporal transformer, embedding size of 16 for each vertex of the graph, a 400-dimensional temporal encoding for the temporal transformer, a Chebyshev polynomial of order 3 for the spatial GCN and 2 heads for the multi-head self-attention for both spatial and temporal transformer. 
\end{itemize}

\subsection{Evaluation Metrics}
In this paper, we use the following three metrics to evaluate the performance of all the models.

\subsubsection{Mean Absolute Error (MAE)}\hfill\\
The Mean Absolute Error (MAE) is defined as:
\begin{equation}
    MAE = \frac{1}{N}\sum_{i} \left( \mid y_t[i] - y_p[i] \mid \right)
    \label{equation:MAE}
\end{equation}

\subsubsection{Root Mean Square Log Error (RMSLE)}\hfill\\
The Root Mean Square Log Error (RMSLE) is defined as:
\begin{equation}
    RMSLE = \sqrt{\frac{1}{N}\sum_{i} \left[ \log (y_t[i]+1) - \log (y_p[i]+1) \right] ^2}
    \label{equation:RMSLE}
\end{equation}

where, $\log$ denotes the natural logarithm.

\subsubsection{Root Mean Square Error (RMSE)}\hfill\\
The Root Mean Square Error (RMSE) is defined as:
\begin{equation}
    RMSE = \sqrt{\frac{1}{N}\sum_{i} \left( y_t[i] - y_p[i] \right) ^2}
    \label{equation:RMSE}
\end{equation}

In all of the above equations, $N$ is the total number of samples in the dataset, $y_t$ is the number of ground truth cases and $y_p$ is the forecasted number of cases. A lower MAE, RMSLE and RMSE indicate that the forecasted number of cases is closer to the ground truth number of cases.

\subsection{Implementation Details}
For all our experiments, we used the past 12 days history (M=12) to forecast the next 12 days horizon (H=12) using the sliding window approach shown in Figure \ref{fig:figure1}. The rationale behind the choice of 12 days history and horizon is that most of the recent state-of-the-art spatial-temporal forecasting algorithms \cite{guo2019attention},\cite{ijcai_GWNET},\cite{yu2018spatio} use the previous 12 time-steps to forecast the next 3, 6, 9 or 12 time-steps. We generate the adjacency matrix $A$ for the spatial graph $G$ at a given time-step by the euclidean distances between the states of the US. This distance is calculated by considering the coordinates of each state which in turn is the average of the latitude and longitude of all the counties for a given state. After generating the adjacency matrix, we normalized it with the maximum value and used a maximum threshold value of $0.3$ to generate the final adjacency matrix for all US states and the counties at the state level. The same adjacency matrix was used for both JHU and NYT datasets to forecast the daily infected and death cases for all 50 US states and Washington, D.C. A separate adjacency matrix was generated at the state level, e.g. the state of Michigan, to forecast the daily infected cases for all the counties.

We divided the entire dataset in chronological order with $80\%$ training, $10\%$ validation and $10\%$ testing. In other words, we used the data from March 15, 2020 to July 28, 2021 as the training set, July 29 to Sept 28, 2021 as the validation set and Sept 29 to Nov 30, 2021 as the testing set. For the NYT dataset, March 18, 2020 was used as the start date as the cases for one of the states were not reported till this date. It should be noted that cross-validation cannot be performed since the validation dataset has to follow the chronological order. Z-score normalization was applied to the inputs, i.e., the train, validation and test datasets were separately normalized by subtracting the mean and dividing by the standard deviation. We used the Mean Absolute Error (MAE) as the loss function and trained all the models for $100$ epochs with early stopping by monitoring the MAE on the validation set. We clipped the gradients to $L2=5$ and used a batch size of 16 for training all models with a learning rate of $0.001$ without using any learning rate decay. 

The performance of our proposed STSGT framework is dependent on several hyper-parameters. These hyper-parameters include the number of STSGT layers, the number of attention heads for multi-head attention in each STST layer, the dimension of the Query(Q), Key(K) and Value(V) matrices for calculating the self-attention within each attention head, and the length of the history time-steps (M) for 12 horizon time-steps (H=12) in the sliding window (see Figure \ref{fig:figure1}). We empirically decided these hyper-parameters such that they are consistent with other spatial-temporal forecasting algorithms and thus we can have a fair comparison. Specifically, we used 2 STSGT layers in our model. Within each STSGT layer, we used 2 STST+GCN layer combinations. For the self-attention, we selected the Query(Q), Key(K) and Value(V) dimensions to be 16 with 2 attention heads in each of the STST layers. Additionally, for the sliding window with 12 horizon time-steps (H=12), we choose 12 historical time-steps (M=12). 

\subsection{Experimental Results}
\subsubsection{Forecasting JHU COVID-19 Daily Infected Cases for US States}\hfill
The results of using our proposed STSGT model and other models to forecast the daily infected cases for the 50 US states and Washington, D.C. using the JHU dataset are summarized in Table \ref{tab:table2}. We summed the daily infected cases for all counties of a given state to generate the ground truth number of infected cases for that state. Clearly, our proposed STSGT significantly outperformed all the other models by achieving the lowest MAE, RMSLE and RMSE for both short-term horizon (next day, $H=1$) and long-term horizon indicated by the mean over the next 12 days ($H=12$). When comparing with the second best model ASTGCN(r), we observe that the mean MAE over the long-term, i.e. 12 days horizon, reduces by 131.22 while the RMSLE and RMSE reduces by 0.081 and 397.81 respectively. An interesting observation is that ASTGCN(r) and our proposed STSGT can better forecast the longer time horizons for all the states of US, compared to other models as seen by a lower 12 days mean MAE when compared to day 1 MAE.

\begin{table}[htbp]
\centering
\caption{Results demonstrating the MAE, RMSLE and RMSE of the daily infected cases of 50 US states and Washington, D.C. with JHU data. The lowest MAE, RMSLE and RMSE are marked in \textbf{bold}.}
\label{tab:table2}
\begin{tabular}{ccccc}
\hline
\hline
Algorithm                                                              & \begin{tabular}[c]{@{}c@{}}Forecasting \\ Horizon\end{tabular} & MAE              & RMSLE          & RMSE             \\ \hline \hline
\multirow{2}{*}{ARIMA \cite{li2018dcrnn_traffic}}                                                 & Day 1 $(H=1)$                                                         & 2204.42          & 5.996          & 3724.19          \\  
                                                                       & 12 Days Mean                                                  & 2206.14          & 6.069          & 3895.20          \\ \hline
\multirow{2}{*}{STGCN \cite{yu2018spatio}}                                                 & Day 1 $(H=1)$                                                         & 1243.48          & 0.869          & 2518.78          \\  
                                                                       & 12 Days Mean                                                  & 1296.86          & 0.905          & 2369.93          \\ \hline
\multirow{2}{*}{ASTGCN(r) \cite{guo2019attention}}                                              & Day 1 $(H=1)$                                                         & 1078.06          & 0.822          & 2369.56          \\  
                                                                       & 12 Days Mean                                                  & 1076.29          & 0.903          & 2102.68          \\ \hline
\multirow{2}{*}{\begin{tabular}[c]{@{}c@{}}Graph\\ WaveNet \cite{ijcai_GWNET}\end{tabular}}                                         & Day 1 $(H=1)$                                                         & 1140.02          & 0.917          & 2203.16          \\  
                                                                       & 12 Days Mean                                                  & 1315.56          & 0.969          & 2263.27          \\ \hline
\multirow{2}{*}{STTN \cite{xu2020spatial}}                                                  & Day 1 $(H=1)$                                                         & 1172.34          & 0.948          & 2475.33          \\ 
                                                                       & 12 Days Mean                                                  & 1198.16          & 0.984          & 2191.91          \\ \hline
\multirow{2}{*}{\begin{tabular}[c]{@{}c@{}}STSGT\\ (ours)\end{tabular}} & Day 1 $(H=1)$                                                         & \textbf{978.390} & \textbf{0.802} & \textbf{2129.74} \\ 
                                                                       & 12 Days Mean                                                  & \textbf{945.070} & \textbf{0.822} & \textbf{1704.87} \\ \hline \hline
\end{tabular}
\end{table}

\begin{figure}[htbp]
\centering
\includegraphics[width=\linewidth]{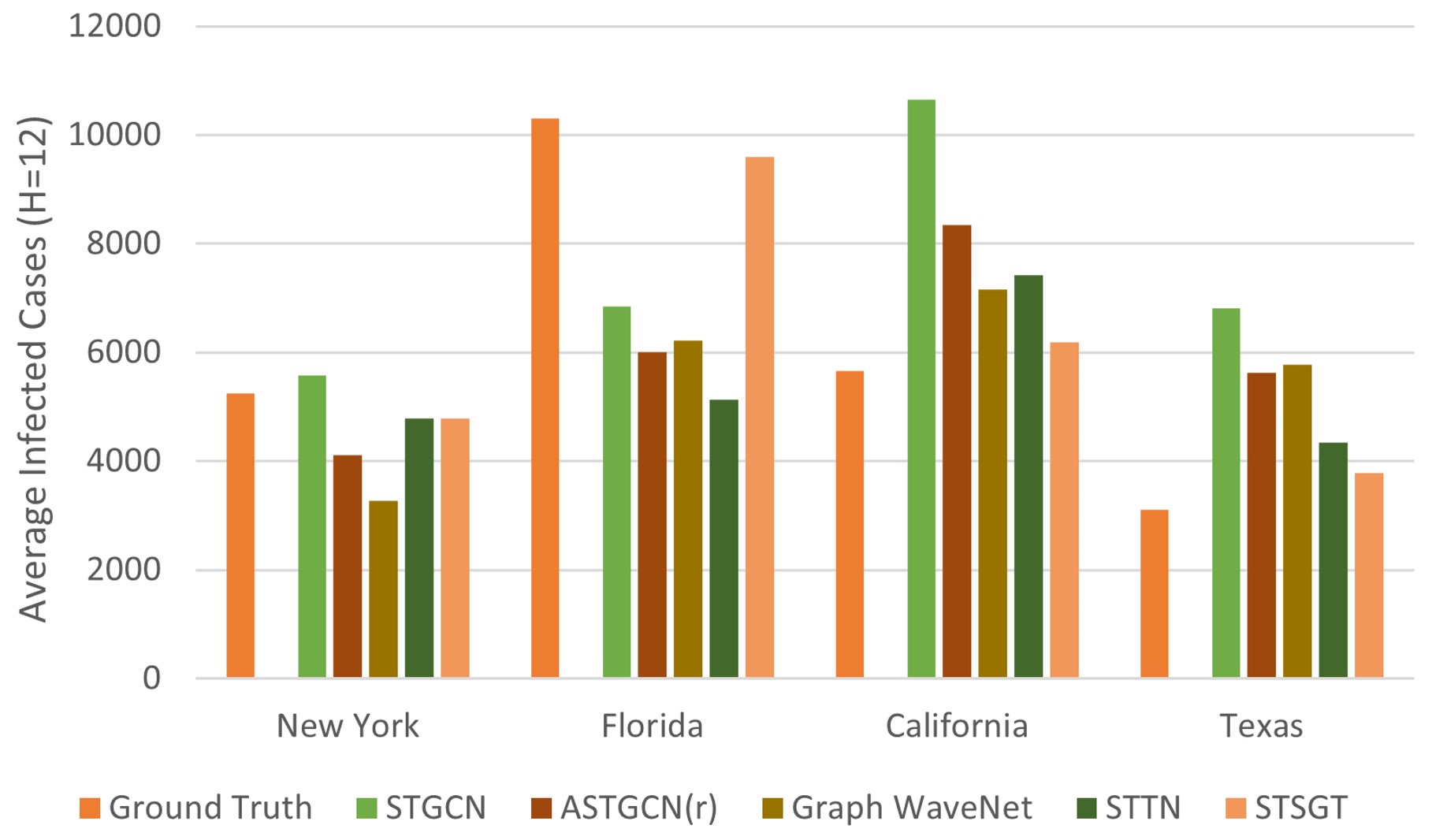}
\caption{A comparison of the average infected cases for the next 12 days horizon $(H=12)$ for four US states.}
\label{fig:figure7}
\end{figure}

A comparison of the average infected cases over the next 12 days horizon $(H=12)$ for four of the most affected US states due to COVID, is shown in Figure \ref{fig:figure7}. We compared the average ground truth and forecasted average infected cases for the states of New York, Florida, California and Texas for a period of 12 days ranging from Nov 6-17, 2021, which is a time frame from our testing set. We use the past 12 days historical data $(M=12)$ ranging from Oct 25-Nov 5, 2021, to forecast the next 12 days horizon $(H=12)$ infected cases (Nov 6-17, 2021). In Figure \ref{fig:figure7}, we observe that the average number of forecasted infected cases with our proposed STSGT is much closer to the ground truth number of average infected cases for most of the states, when compared to state-of-the-art spatial-temporal forecasting methods.

\subsubsection{Forecasting JHU COVID-19 Daily Death Cases for US States}\hfill
The results of using our proposed STSGT model and other models to forecast the daily death cases for the 50 US states and Washington, D.C. using the JHU dataset are summarized in Table \ref{tab:table3}. Similar to Table \ref{tab:table2}, we summed the daily death cases for all counties of a given state to generate the ground truth number of death cases for that state. Our proposed STSGT outperformed all the other models by achieving the lowest MAE for the short-term horizon (H=1) and the lowest MAE, RMSLE and RMSE for the long-term horizon indicated by the mean over the next 12 days (H=12). When comparing with the second best model ASTGCN(r), we observe that the mean MAE over the long-term, i.e. 12 days horizon, reduces by 0.73 while the RMSLE and RMSE reduces by 0.005 and 0.11 respectively. We observe that even though ASTGCN(r) achieves lower RMSLE and RMSE for forecasting the next day (day 1) death cases, our proposed model STSGT could better forecast the death cases over the long-term (12 days) as observed by a lower MAE, RMSLE and RMSE.

\begin{table}[htbp]
\centering
\caption{Results demonstrating the MAE, RMSLE and RMSE of the daily death cases of 50 US states and Washington, D.C. with JHU data. The lowest MAE, RMSLE and RMSE are marked in \textbf{bold}.}
\label{tab:table3}
\begin{tabular}{ccccc}
\hline
\hline
Algorithm                                                              & \begin{tabular}[c]{@{}c@{}}Forecasting \\ Horizon\end{tabular} & MAE            & RMSLE          & RMSE           \\ \hline \hline
\multirow{2}{*}{ARIMA \cite{li2018dcrnn_traffic}}                                                 & Day 1 $(H=1)$                                                         & 42.62          & 2.753          & 74.89          \\  
                                                                       & 12 Days Mean                                                  & 43.66          & 2.698          & 110.60         \\ \hline
\multirow{2}{*}{STGCN \cite{yu2018spatio}}                                                 & Day 1 $(H=1)$                                                         & 26.25          & 0.966          & 55.58          \\  
                                                                       & 12 Days Mean                                                  & 25.23          & 0.952          & 54.15          \\ \hline
\multirow{2}{*}{ASTGCN(r) \cite{guo2019attention}}                                              & Day 1 $(H=1)$                                                         & 21.20          & \textbf{0.849} & \textbf{47.73} \\ 
                                                                       & 12 Days Mean                                                  & 21.36          & 0.873          & 47.25          \\ \hline
\multirow{2}{*}{\begin{tabular}[c]{@{}c@{}}Graph\\ WaveNet \cite{ijcai_GWNET}\end{tabular}}                                         & Day 1 $(H=1)$                                                         & 23.02          & 0.889          & 48.99          \\ 
                                                                       & 12 Days Mean                                                  & 25.03          & 0.922          & 51.16          \\ \hline
\multirow{2}{*}{STTN \cite{xu2020spatial}}                                                  & Day 1 $(H=1)$                                                         & 21.31          & 0.863          & 48.52          \\ 
                                                                       & 12 Days Mean                                                  & 22.05          & 0.892          & 48.23          \\ \hline
\multirow{2}{*}{\begin{tabular}[c]{@{}c@{}}STSGT\\ (ours)\end{tabular}} & Day 1 $(H=1)$                                                         & \textbf{20.73} & 0.855          & 48.30          \\ 
                                                                       & 12 Days Mean                                                  & \textbf{20.63} & \textbf{0.868} & \textbf{47.14} \\ \hline \hline
\end{tabular}
\end{table}

\subsubsection{Forecasting JHU COVID-19 Daily Infected Cases for Michigan}
The results of using our proposed STSGT model and other models to forecast the daily infected cases at the state level, e.g., the state of Michigan, using the JHU dataset, are summarized in Table \ref{tab:table4}. In this case, the daily infected cases of the individual counties of Michigan served as the ground truth. We excluded the correctional facilities, counties denoted as ``Out of Michigan'' and unassigned counties, thus resulting in 83 counties in total. Hence, we used 83 vertices for the graph per time-step (i.e. for each day). As shown in Table \ref{tab:table4}, our proposed STSGT significantly outperformed all the other models by achieving the lowest MAE, RMSLE and RMSE for both short-term horizon (H=1) and long-term horizon indicated by the mean over the next 12 days (H=12). When comparing with the second best model STTN, we observe that the mean MAE over the long-term horizon (12 days) reduces by 9.86 while the RMSLE and RMSE reduces by 0.267 and 17.49 respectively. This experiment clearly demonstrates that our model could perform very well at the state level and hence our model could be used for forecasting infected and death cases for any US state.

\begin{table}[htbp]
\centering
\caption{Results demonstrating the MAE, RMSLE and RMSE of the daily infected cases for 83 counties of the state of Michigan with JHU data. The lowest MAE, RMSLE and RMSE are marked in \textbf{bold}.}
\label{tab:table4}
\begin{tabular}{ccccc}
\hline
\hline
Algorithm                                                              & \begin{tabular}[c]{@{}c@{}}Forecasting \\ Horizon\end{tabular} & MAE            & RMSLE          & RMSE           \\ \hline \hline
\multirow{2}{*}{ARIMA \cite{li2018dcrnn_traffic}}                                                 & Day 1 $(H=1)$                                                         & 124.52         & 3.851          & 270.78         \\ 
                                                                       & 12 Days Mean                                                  & 142.46         & 4.006          & 322.06         \\ \hline
\multirow{2}{*}{STGCN \cite{yu2018spatio}}                                                 & Day 1 $(H=1)$                                                         & 48.30          & 0.739          & 95.99          \\ 
                                                                       & 12 Days Mean                                                  & 57.31          & 0.753          & 126.73         \\ \hline
\multirow{2}{*}{ASTGCN(r) \cite{guo2019attention}}                                              & Day 1 $(H=1)$                                                         & 47.50          & 0.772          & 97.14          \\ 
                                                                       & 12 Days Mean                                                  & 55.20          & 0.785          & 122.10         \\ \hline
\multirow{2}{*}{\begin{tabular}[c]{@{}c@{}}Graph\\ WaveNet \cite{ijcai_GWNET}\end{tabular}}                                          & Day 1 $(H=1)$                                                         & 41.34          & 0.658          & 87.97          \\ 
                                                                       & 12 Days Mean                                                  & 64.05          & 0.879          & 143.61         \\ \hline
\multirow{2}{*}{STTN \cite{xu2020spatial}}                                                  & Day 1 $(H=1)$                                                         & 42.50          & 0.812          & 81.37          \\  
                                                                       & 12 Days Mean                                                  & 49.33          & 0.810          & 108.31         \\ \hline
\multirow{2}{*}{\begin{tabular}[c]{@{}c@{}}STSGT\\ (ours)\end{tabular}} & Day 1 $(H=1)$                                                         & \textbf{22.04} & \textbf{0.430} & \textbf{40.78} \\  
                                                                       & 12 Days Mean                                                  & \textbf{39.47} & \textbf{0.543} & \textbf{90.82} \\ \hline \hline
\end{tabular}
\end{table}

\begin{figure}[htbp]
\centering
\includegraphics[width=\linewidth]{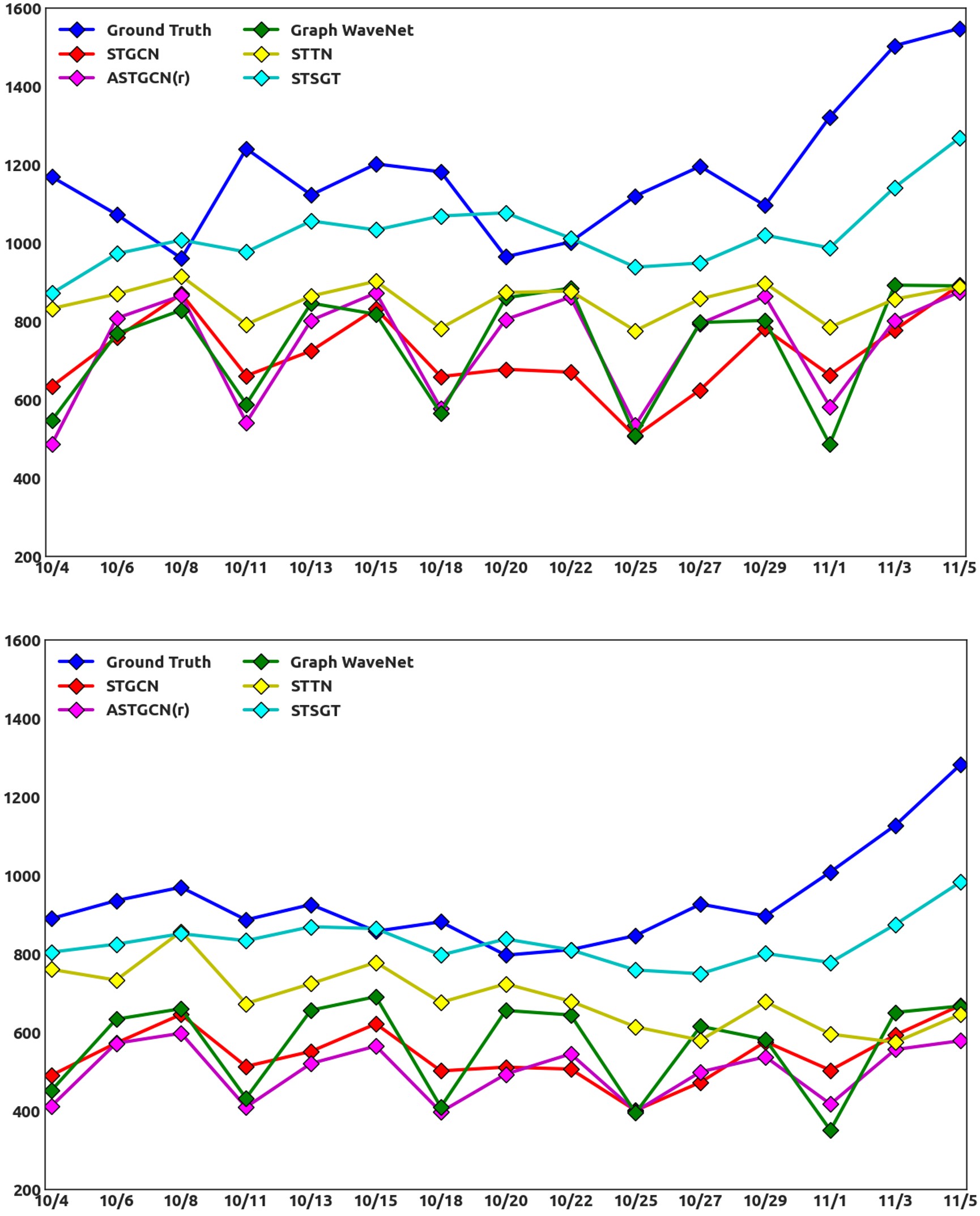}
\caption{(top) Daily Infected cases for Wayne County, Michigan and (bottom) Daily Infected cases for Oakland County, Michigan from October 4 to November 5, 2021.}
\label{fig:figure8}
\end{figure}

The two most populated counties in Michigan are Wayne county with a population over 1.7 million and Oakland county with a population over 1.2 million. We plotted the ground truth and forecasted number of infected cases for Wayne and Oakland counties for a time frame in the testing set ranging from Oct 4-Nov 5, 2021 as shown in Figure \ref{fig:figure8}. We use the 12 days historical data $(M=12)$ to forecast the next day infected cases $(H=1)$ in Figure \ref{fig:figure8}. Specifically, we use the data from Sept 22-Oct 3 to forecast the infected cases for Oct 4, then the data from Sept 24-Oct 5 to forecast the infected cases for Oct 6 and so on. We removed the days when the ground truth infected cases were zero as these cases were not reported in the JHU dataset for Michigan. In Figure \ref{fig:figure8}, we compare STSGT with other spatial-temporal forecasting algorithms and clearly observe that the forecasted number of infected cases with STSGT is much closer to the ground truth number of infected cases. 

\subsubsection{Forecasting NYT COVID-19 Daily Infected Cases for US States}\hfill
The results of using our proposed STSGT model and other models to forecast the daily infected cases for the 50 US states and Washington, D.C. using the NYT dataset are summarized in Table \ref{tab:table5}. For the NYT dataset, the daily infected cases were reported for each state thereby eliminating the need for the county-wise summation process as done for the JHU dataset. Clearly, our proposed STSGT significantly outperformed all the other models by achieving the lowest MAE and RMSE for the short-term horizon (H=1) and the lowest MAE, RMSLE and RMSE for the long-term horizon indicated by the mean over the next 12 days (H=12).
When comparing with the second best model ASTGCN(r), we observe that the mean MAE over the long-term horizon (12 days) reduces by 52.77 while the RMSLE and RMSE reduces by 0.010 and 136.08 respectively. This clearly shows the advantage of our model for forecasting over the long-term horizon.  

\begin{table}[htbp]
\centering
\caption{Results demonstrating the MAE, RMSLE and RMSE of the daily infected cases of 50 US states and Washington, D.C. with NYT data. The lowest MAE, RMSLE and RMSE are marked in \textbf{bold}.}
\label{tab:table5}
\begin{tabular}{ccccc}
\hline
\hline
Algorithm                                                              & \begin{tabular}[c]{@{}c@{}}Forecasting \\ Horizon\end{tabular} & MAE              & RMSLE          & RMSE             \\ \hline \hline
\multirow{2}{*}{ARIMA \cite{li2018dcrnn_traffic}}                                                 & Day 1 $(H=1)$                                                         & 2152.48          & 5.935          & 3311.26          \\  
                                                                       & 12 Days Mean                                                  & 2185.04          & 5.968          & 3676.43          \\ \hline
\multirow{2}{*}{STGCN \cite{yu2018spatio}}                                                 & Day 1 $(H=1)$                                                         & 1340.13          & 0.975          & 2303.10          \\ 
                                                                       & 12 Days Mean                                                  & 1384.35          & 0.997          & 2357.91          \\ \hline
\multirow{2}{*}{ASTGCN(r) \cite{guo2019attention}}                                              & Day 1 $(H=1)$                                                         & 939.840           & \textbf{0.852} & 1582.46          \\ 
                                                                       & 12 Days Mean                                                  & 976.420           & 0.879          & 1626.42          \\ \hline
\multirow{2}{*}{\begin{tabular}[c]{@{}c@{}}Graph\\ WaveNet \cite{ijcai_GWNET}\end{tabular}}                                         & Day 1 $(H=1)$                                                         & 960.840           & 0.854          & 1506.59          \\  
                                                                       & 12 Days Mean                                                  & 1173.00          & 0.919          & 1905.55          \\ \hline
\multirow{2}{*}{STTN \cite{xu2020spatial}}                                                  & Day 1 $(H=1)$                                                         & 1125.88          & 0.978          & 1794.09          \\ 
                                                                       & 12 Days Mean                                                  & 1177.85          & 1.029          & 1854.19          \\ \hline
\multirow{2}{*}{\begin{tabular}[c]{@{}c@{}}STSGT\\ (ours)\end{tabular}} & Day 1 $(H=1)$                                                         & \textbf{903.590} & 0.864          & \textbf{1465.55} \\ 
                                                                       & 12 Days Mean                                                  & \textbf{923.650} & \textbf{0.869} & \textbf{1490.34} \\ \hline \hline
\end{tabular}
\end{table}

\subsubsection{Forecasting NYT COVID-19 Daily Death Cases for US States}\hfill
The results of using our proposed STSGT model and other models to forecast the daily death cases for the 50 US states and Washington, D.C. using the NYT dataset are summarized in Table \ref{tab:table6}. Similar to Table \ref{tab:table5}, the daily death cases were reported for each state thereby eliminating the need for the county-wise summation process as done for the JHU dataset. Clearly, our proposed STSGT significantly outperformed all the other models by achieving the lowest MAE and RMSLE for the short-term horizon ($H=1$) and the lowest MAE, RMSLE and RMSE for the long-term horizon indicated by the mean over the next 12 days ($H=12$). When comparing with the second best model ASTGCN(r) for the long-term (12 days mean), we observe that the mean MAE over 12 days horizon reduces by 2.02 while the RMSLE and RMSE reduces by 0.025 and 0.02 respectively. 

\begin{table}[htbp]
\centering
\caption{Results demonstrating the MAE, RMSLE and RMSE of the daily death cases of 50 US states and Washington, D.C. with NYT data. The lowest MAE, RMSLE and RMSE are marked in \textbf{bold}.}
\label{tab:table6}
\begin{tabular}{ccccc}
\hline
\hline
Algorithm                                                              & \begin{tabular}[c]{@{}c@{}}Forecasting \\ Horizon\end{tabular} & MAE            & RMSLE          & RMSE           \\ \hline \hline
\multirow{2}{*}{ARIMA \cite{li2018dcrnn_traffic}}                                                 & Day 1 $(H=1)$                                                         & 49.25          & 2.693          & 107.41         \\ 
                                                                       & 12 Days Mean                                                  & 49.36          & 2.661          & 138.96         \\ \hline
\multirow{2}{*}{STGCN \cite{yu2018spatio}}                                                 & Day 1 $(H=1)$                                                         & 28.86          & 0.949          & 67.85          \\  
                                                                       & 12 Days Mean                                                  & 28.43          & 0.959          & 63.87          \\ \hline
\multirow{2}{*}{ASTGCN(r) \cite{guo2019attention}}                                              & Day 1 $(H=1)$                                                         & 25.90          & 0.944          & 62.07          \\  
                                                                       & 12 Days Mean                                                  & 24.40          & 0.932          & 57.47          \\ \hline
\multirow{2}{*}{\begin{tabular}[c]{@{}c@{}}Graph\\ WaveNet \cite{ijcai_GWNET}\end{tabular}}                                         & Day 1 $(H=1)$                                                         & 24.73          & 0.921          & \textbf{60.07} \\  
                                                                       & 12 Days Mean                                                  & 26.18          & 0.965          & 60.57          \\ \hline
\multirow{2}{*}{STTN \cite{xu2020spatial}}                                                  & Day 1 $(H=1)$                                                         & 29.85          & 1.091          & 69.55          \\  
                                                                       & 12 Days Mean                                                  & 28.09          & 1.079          & 62.97          \\ \hline
\multirow{2}{*}{\begin{tabular}[c]{@{}c@{}}STSGT\\ (ours)\end{tabular}} & Day 1 $(H=1)$                                                         & \textbf{22.88} & \textbf{0.912} & 61.39          \\ 
                                                                       & 12 Days Mean                                                  & \textbf{22.38} & \textbf{0.907} & \textbf{57.45} \\ \hline \hline
\end{tabular}
\end{table}

\section{Discussion and Future Work}
The present study investigated the impact and utility of spatial-temporal forecasting for COVID-19 time series analysis. Our proposed novel STSGT could accurately forecast the daily infected and death cases across the 50 US states and Washington, D.C. STSGT also performed extremely well at the state level by accurately forecasting the daily infected cases for all the counties of Michigan. Hence, our model could help the common people and the government by taking necessary precautions so that COVID could be prevented. 

It should be noted that we have selected mid-March (March 15 or 18, 2020) as the start date for our analysis as the pandemic started in almost all of the US states around that date, and there were significant number of infected and death cases after mid-March. We chose Nov 30, 2021 as the end date of our analysis as the Omicron variant of COVID started dominating after this date, and the number of daily cases spiked to a new high. Hence, the daily cases from around December 1, 2021 until the end of Omicron wave (around Feb 15, 2022) formed a different pattern than what we observed from the start of the pandemic. We believe that our model needs to be re-trained again with the Omicron wave's new data pattern to achieve accurate forecasting, and we leave this for future work. In future, we also plan to incorporate an efficient self-attention mechanism to further accelerate the training process of STSGT while maintaining similar performance as our current approach.

\section{Conclusion}
In this paper, we proposed a novel model which can synchronously capture the complex spatial-temporal dependencies in the COVID data using the multi-head self-attention. We show through extensive experiments on two real-world public datasets that our model can outperform the existing models specifically designed for spatial-temporal forecasting. Our proposed model could have a significant impact on society by making the common people more aware and help the government eradicate this pandemic in the near future.
\begin{acks}
This work was partially supported by National Science Foundation (Grant No. 2027251).
\end{acks}
\bibliographystyle{ACM-Reference-Format}
\bibliography{references}

\end{document}